%% file: acl_latex.tex
\pdfoutput=1
\documentclass[11pt]{article}

\usepackage[preprint]{acl}

\usepackage{times}
\usepackage{latexsym}

\usepackage{xspace}
\usepackage{url}
\usepackage[most]{tcolorbox}
\usepackage{subcaption}
\usepackage{soul}
\usepackage{hyperref}
\usepackage{titlesec}
\usepackage{lipsum}% provides filler text
\titlespacing{\paragraph}{%
  0pt}{%              left margin
  0.5\baselineskip}{% space before (vertical)
  1em}%               space after (horizontal)
\usepackage[most]{tcolorbox}
\usepackage{colortbl} 
\usepackage{arydshln}
\usepackage[font={small}]{caption}
\usepackage{multirow}
\usepackage{cleveref}
\usepackage{fontawesome5} % for the GitHub icon

\crefname{section}{\S}{\S}

\PassOptionsToPackage{dvipsnames,xcdraw,table}{xcolor}

\usepackage[T1]{fontenc}
\usepackage[utf8]{inputenc}

\usepackage{microtype}
\usepackage{enumitem}

\usepackage{inconsolata}

\usepackage{graphicx}

\title{Evaluating the Evaluators:\\Are readability metrics good measures of readability?}

\author{Isabel Cachola, \; Daniel Khashabi$^*$ \textnormal{ and } Mark Dredze$^*$ \\
        Johns Hopkins University, Baltimore, MD 21211 \\
        \texttt{\{icachola, danielk, mdredze\}@cs.jhu.edu}
        }

\makeatletter
\def\blfootnote{\xdef\@thefnmark{}\@footnotetext}
\makeatother

\begin{document}
\include{commands}
\maketitle
\begin{abstract}

    Plain Language Summarization (PLS) aims to distill complex documents into accessible summaries for non-expert audiences. In this paper, we conduct  a thorough survey of PLS literature, and identify that the current standard practice for readability evaluation is to use traditional readability metrics, such as Flesch-Kincaid Grade Level (FKGL). However, despite proven utility in other fields, these metrics have not been compared to human readability judgments in PLS.  We evaluate 8 readability metrics and show that most correlate poorly with human judgments, including the most popular metric, FKGL. We then show that Language Models (LMs) are better judges of readability, with the best-performing model achieving a Pearson correlation of $0.56$ with human judgments. Extending our analysis to PLS datasets, which contain summaries aimed at non-expert audiences, we find that LMs better capture deeper measures of readability, such as required background knowledge, and lead to different conclusions than the traditional metrics. Based on these findings, we offer recommendations for best practices in the evaluation of plain language summaries. We release our analysis code and survey data.  
    \bigskip\\
    \href{https://github.com/JHU-CLSP/eval-the-eval-readability}{\faGithub\xspace
    \texttt{JHU-CLSP/eval-the-eval-readability}}

\end{abstract}

\blfootnote{\hspace{-0.23cm} $^*$Equal advising.}

\input{1-Introduction}

\input{2-Related-Works}
\input{3-Experimental-Setup}

\input{4-Results}
\input{5-Discussion}

\section*{Limitations}
The conclusions of this paper are limited to the task of plain language summarization, and are not intended to apply to other applications of readability metrics, such as judging the age-level appropriateness of educational material. Additionally, our human judgments and experiments focused on the summarization of scientific articles, and may not generalize to PLS in other domains, such as law or clinical notes. Finally, our experiments are limited to the English language, and our findings may not apply to other languages. We leave the exploration of readability evaluation in other domains and languages to future work.

\section*{Ethical Considerations}
This paper involves the use of LMs for generation and evaluation. LMs have been shown to generate factually incorrect information and are subject to bias~\cite{Venkit2024AnAO,Stureborg2024LargeLM}. Additionally, the use of language models contributes to the environmental footprint of our field~\cite{Schwartz2019GreenA}. However, this paper focuses on the evaluation of plain language summarization, which has the potential to make scientific knowledge more accessible to the general population. Therefore, we believe that the benefits of this work outweigh the potential harms.

\section*{Acknowledgments}
We'd like to thank the authors of \citet{August2024KnowYA} for sharing their human-annotated dataset with us. We'd additionally like to thank Tal August for his insightful guidance in creating this paper. 
DK was supported by ONR (N00014-24-1-2089).

\bibliography{custom,daniel_ref}
\clearpage
\input{appendix}
\end{document}

%% file: commands.tex
\newcommand\isabel[1]{{\color{purple}\{#1\}$_{Isabel}$}}

\newcommand\todoit[1]{{\color{red}\{TODO: \textit{#1}\}}}
\newcommand\todo{{\color{red}{TODO}}\xspace}
\newcommand\X{{\color{red}{X}}\xspace}
\newcommand\todocite{{\color{red}{CITE}}\xspace}
\newcommand{\red}[1]{\textcolor{red}{#1}} 
\newcommand{\blue}[1]{\textcolor{blue}{#1}} 

\newtcbox{\badge}[1][red]{
  on line, 
  arc=2pt,
  colback=#1!60!black,
  colframe=#1!60!black,
  fontupper=\color{white},
  boxrule=1pt, 
  boxsep=0pt,
  left=2pt,
  right=2pt,
  top=1pt,
  bottom=1pt
}

\newcommand{\textexample}[1]{
    \begin{tcolorbox}[
                  colback=gray!20,%gray background
                  colframe=gray,
                  boxsep=0pt,
                  left=2pt,
                  right=2pt,
                  top=2pt,
                  bottom=2pt,
                 ]{\footnotesize#1}
                 \end{tcolorbox}
}

\newcommand{\rqone}{\small \badge[red]{RQ1}\xspace}
\newcommand{\rqtwo}{\small \badge[orange]{RQ2}\xspace}
\newcommand{\rqthree}{\small \badge[green]{RQ3}\xspace}
\newcommand{\rqfour}{\small \badge[blue]{RQ4}\xspace}

\newcommand{\hdash}{\hdashline[0.5pt/3pt]}
\newcommand{\textnewline}{\xspace\textbackslash n}

%% file: 1-Introduction.tex
\section{Introduction}

In the field of Natural Language Processing (NLP), plain language summarization (PLS) distills complex documents, such as scientific articles, into accessible summaries for non-expert audiences while preserving essential meaning~\cite{chandrasekaran-etal-2020-overview-insights}. The COVID-19 pandemic highlighted the critical need to make scientific knowledge accessible to the general public~\cite{wang-etal-2020-cord}. By enhancing public engagement with research, PLS can help bridge the gap between expert knowledge and general understanding.

Although human evaluation remains the gold standard for assessing summary quality and readability, the high cost and slow turnaround~\cite{Liu2022RevisitingTG} have led many researchers to rely on automatic evaluation metrics for evaluating PLS summaries~\cite{Goldsack2022MakingSS, Guo2020AutomatedLL}. Although these metrics have been validated in fields such as education and law~\cite{ef903df0-98e8-341d-9e07-ada8904086a0, Han2024TheUO}, their effectiveness in reflecting readability in the context of PLS remains unproven.

Are automated readability metrics appropriate evaluators for the task of PLS? We explore whether the definition of readability as implemented by automated measures matches the definition used by the PLS research community. Additionally, given that Language Models (LMs) can reason over complex language tasks~\cite{Brown2020LanguageMA, Wei2022ChainOT, Yang2024DoLL}, we explore whether LMs can judge the readability of a summary. Given these motivations, we ask the following research questions (RQs).

\paragraph{{\rqone\label{rq1}}What is the current standard of evaluation in PLS literature?} 
We review PLS literature by collecting relevant papers published in *ACL venues from 2013 to 2025 and note the readability evaluation method used in the study. We find that the majority of papers focus on a small number of traditional readability metrics, such as Flesch-Kincaid grade Level (FKGL)~\cite{flesch1952simplification}.
This finding motivates our analysis of the suitability of traditional readability metrics for PLS evaluation.

\paragraph{{\rqtwo\label{rq2}}How well do traditional readability metrics correlate with human readability judgments?} Since the PLS research community employs these traditional metrics (RQ1), we assess their suitability by measuring their correlation with human readability judgments. A low correlation would suggest that a metric is inadequate for evaluating PLS readability, and would necessitate the PLS research community identify and move to better metrics. To the best of our knowledge, this work is the first to compare traditional readability metrics to human readability judgments for PLS.

\paragraph{{\rqthree\label{rq3}}How well do LM-based evaluators correlate with human readability judgments?} Traditional readability metrics primarily use lexical features, such as the number of syllables in a word, to measure readability. In contrast, LMs may capture more complex attributes of readability than traditional metrics, such as the inclusion of necessary context and explanation of key concepts. The findings of this research question have important implications for both the best practices in evaluation of PLS and the broader NLP community's understanding of LM capabilities.

\paragraph{{\rqfour\label{rq4}} What do LM-based evaluators reveal about the readability of popular summarization datasets?} Researchers often rely on traditional readability metrics when assessing summaries in new methods or datasets. However, if these metrics correlate poorly with human judgments, the resulting conclusions may be flawed. Similarly, existing datasets, which often arise from data of convenience, may be poorly suited to PLS research. This RQ explores what LM-based evaluators reveal about the readability of popular summarization datasets and how LM-based conclusions differ from those based on traditional readability metrics. 

We answer these questions through the following contributions. First, we survey PLS papers published in *ACL venues and find that the most popular metric for readability evaluation is Flesch-Kincaid Grade Level (FKGL)~\cite{flesch1952simplification}. Motivated by these findings, we then compare 8 traditional readability metrics to human judgments. We show that 6 of the 8 metrics have a poor correlation (less than 0.3 Pearson correlation) with human judgments, including FKGL, indicating that these metrics are poor measures of readability for PLS. Additionally, we compare the judgments of 5 LMs to human judgments and show that LMs outperform the traditional metrics. We demonstrate that LMs have promising potential as evaluators by reasoning over more complex attributes of readability. We use LM evaluators to re-evaluate 10 summarization datasets and show that some summarization datasets intended for PLS achieve similar readability scores to those aimed at expert audiences, calling into question the utility of these data. Finally, based on a thorough analysis of current readability evaluation practices, we offer recommendations for best practices in PLS evaluation and identify opportunities for future work.

	% 5.	Expand the Introduction for a Slower, More Detailed Development
	% •	The introduction should be longer, ideally extending into a second column on the next page, to ensure a thorough setup of the research problem and motivation.
	% 6.	Explicitly Write Out Research Questions with Explanations
	% •	Each research question should be clearly stated, followed by a brief explanation:
	% •	RQ1: How well do traditional readability metrics correlate with human judgments?
	% •	If they do not correlate, then they are not suited for evaluating the success of plain language summarization.
	% 7.	Clarify the Description of Tasks
	% •	When referring to “the nine tasks,” provide a brief explanation of their nature. For example:
	% •	These tasks span various scientific domains, with content aimed at different audiences, such as children or journalists

%% file: 2-Related-Works.tex
\section{Related Works}

% \daniel{consider adding paragraph heading so that the reader can expect what each paragraph is about.}

\paragraph{Summarization evaluation.} PLS research often introduces either datasets~\cite{Goldsack2022MakingSS, Crossley2021TheCE, pu-etal-2024-scinews, Manor2019PlainES}, methods~\cite{Guo2022RetrievalAO, August2022GeneratingSD, Luo2022ReadabilityCB, ji-etal-2024-rag, flores-etal-2023-medical}, or both~\cite{Guo2020AutomatedLL,Zaman2020HTSSAN, chandrasekaran-etal-2020-overview-insights}. The majority of prior work use a combination of readability metrics, such as Flesch Reading Ease~\cite{Flesch1943MarksOR} or the Gunning-Fog Index~\cite{Gunning1968TheTO} to validate the readability of their dataset or generations. Readability metrics are typically reported in conjunction with more general summarization metrics, such as ROUGE~\cite{Lin2004ROUGEAP} or BertScore~\cite{Zhang2019BERTScoreET}. General summarization evaluation is a well-studied area, with ongoing work analyzing both the efficacy of summarization metrics~\cite{Fabbri2020SummEvalRS, khashabi2022genie, Goyal2022NewsSA} and designing metrics that better align with human judgments~\cite{liu-etal-2023-towards-interpretable,Liu2022RevisitingTG}. \citet{Guo2023APPLSEE} analyzed how perturbations in plain language summaries affect results of general summarization metrics. In this work, we focus on readability metrics, rather than general summarization metrics, with the goal of understanding how well readability metrics measure readability for PLS.

\paragraph{Readability Metrics.} While readability metrics are well studied in fields such as education~\cite{ef903df0-98e8-341d-9e07-ada8904086a0, dubay2004principles, texts_2022} and linguistics~\cite{Pires02102017}, there is little work studying how well these metrics perform for the task of PLS. 
Most traditional metrics were not designed specifically for PLS, or even for evaluation in Computer Science. The most common origin of traditional metrics is the need to assess the readability of K-12 school texts~\cite{dale1948formula, Coleman1975ACR}. %Flesch Reading Ease was developed for the U.S. Navy to asses the difficulty of technical manuals~\cite{Flesch1943MarksOR}. 
Linsear Write was introduced in the book, \textit{Gobbledygook has gotta go}, published by the US Department of the Interior for the purposes of measuring the complexity of government communications~\cite{o1966gobbledygook}.
% Coleman-Liau was published for the purposes of rating the readability of public school textbooks on a large scale, with the specific benefit of only needing a an optical scanning device to run the calculations~\cite{Coleman1975ACR}. %In contrast, the majority of PLS literature focuses on generating summaries of scientific papers for a general audience, in which ``general audience'' is defined as an adult who is not an expert in the paper's domain~\todocite. The inherent difference in audience (K-12 vs general adult population) and intent (school-grade education vs scientific communication), means that these metrics may not cleanly apply to the task of PLS. In this paper, we attempt to answer the question of application of readability metrics to PLS.
As readability metrics rely primarily on lexical features~\cite{508dd5a7-e076-39ee-bf49-06f80e20c28d}, prior work has offered criticism of readability metrics, showing that they can be easily manipulated to provide better scores with changes that do not substantially improve the readability of summaries~\cite{Tanprasert2021FleschKincaidIN}. Other work has looked at which linguistic attributes are correlated with readability metrics~\cite{tajner2012WhatCR}. To the best of our knowledge, our work is the first to measure the correlation of readability metrics with human readability judgments.

\paragraph{LMs as Evaluators.} Recent advances in LMs have shown that they are capable of reasoning over complex language~\cite{Brown2020LanguageMA, Wei2022ChainOT, Yang2024DoLL}. LMs have been shown to be effective evaluators in other natural language tasks~\cite{li-etal-2025-exploring-reliability,10.1145/3640457.3688075, nedelchev-etal-2020-language, Liu2023GEvalNE}, including related summarization tasks~\cite{song-etal-2024-finesure}. Given this success in prior work, we hypothesize that LMs are capable of evaluating the readability of plain language summaries. In particular, we hypothesize that LMs can reason over more complex attributes of readability, such as the background required or whether technical concepts are explained. 

%% file: 3-Experimental-Setup.tex
\section{Experimental Setup}\label{sec:experimental-setup}
% In this section, we describe the experimental setup for our three research questions.
\subsection{Current PLS evaluation standards \hyperref[rq1]{\rqone}}

We aim to conduct a thorough literature survey of the standard practices in readability evaluation for PLS. We collect papers\footnote{On May 7th, 2025} from the ACL Anthology\footnote{\url{https://aclanthology.org/}} that mention one of the following key phrases: ``plain language summarization,'' ``readable summaries,'' or ``lay summarization.'' We exclude papers published for a shared task from annotation and assume the participants use the metrics designated by the shared task organizers. Our goal is to understand the decisions made by researchers, and including shared task papers in this survey would over-represent the decisions made by the task organizers. We report the evaluation methods used by the shared tasks and the number of participants to represent the impact of the evaluation choices. We identify 55 papers that match our criteria. We annotate the papers for relevance to PLS, the type of publication (Main conference, Findings, or Workshop), and which readability evaluation metrics are used. We exclude papers from the survey not relevant to PLS, resulting in 18 relevant papers from the years 2013 to 2025. The most common reasons for relevance exclusion include using ``readable'' in a different word sense (e.g. ``human readable'' vs ``machine readable'') or just citing a PLS paper. We report the number of papers that use each  metric.

\subsection{Comparing traditional readability metrics to human judgments \hyperref[rq2]{\rqtwo}}

\paragraph{Human Annotated Data.} 
To measure the correlation between readability metrics with human judgments, we use the dataset collected by ~\citet{August2024KnowYA}. This dataset contains 60 summaries of 10 scientific papers in a variety of domains. Each paper has both expert written and machine written summaries (generated using GPT-3.) The summaries are annotated on a scale of 1 to 5 for the annotator's reading ease of the article. 1 indicates a very poor reading ease, while 5 indicates a very high reading ease. For each summary, we take the average of the annotators' scores to calculate the correlations with readability evaluations as described below. \citet{August2024KnowYA} originally collected this dataset to better understand human preferences in scientific summarization. In this paper, we extend their work by applying their findings to summarization evaluation metrics. To the best of our knowledge, this is the only available dataset of human judgments for PLS. \Cref{appendix:dataset_details} contains additional dataset details.

\paragraph{Traditional readability metrics.} We consider ``traditional'' readability metrics to be those most commonly used in PLS literature. These metrics are well-established, and have been used in past work as judges of readability~\cite{chandrasekaran-etal-2020-overview-insights}. This term excludes LM-based evaluations, discussed in~\Cref{sec:rq-three-setup}.
We consider 8 readability metrics: Flesh-Kincaid Grade Level (FKGL)~\cite{flesch1952simplification}, Flesch Reading Ease (FRE)~\cite{Flesch1943MarksOR}, Dale Chall Readability Score (DCRS)~\cite{dale1948formula}, Automated Readability Index (ARI)~\cite{smith1967automated}, Coleman Liau Index (CLI)~\cite{Coleman1975ACR}, Gunning Fog Index (GFI)~\cite{Isnaeni_2017}, Spache~\cite{spache1953new} and Linsear Write (LW)~\cite{o1966gobbledygook}.
% \daniel{if these have citations, cite here}. 
All of the metrics, except for DCRS and Spache, use lexical features such as number of syllables or length of sentences to measure readability. DCRS and Spache use word familiarity to measure readability, assuming that more common words are easier to read~\cite{dale1948formula,o1966gobbledygook}.%\daniel{either you need to define what familiarity is or you need to cite; otherwise this sentence is incomplete}. 
\footnote{We use the \href{https://github.com/cdimascio/py-readability-metrics}{py-readability-metrics} package to calculate the readability scores.} %All of the metrics listed above, except LW and FRE, provide a lower score for higher readability, while the human judgments provide a higher score for higher readability. 

\paragraph{Quantifying alignment between traditional metrics and humans.} We report the Pearson and Kendall-Tau correlation of each metric listed above with the human judgments collected by~\citet{August2024KnowYA}.
Except for LW and FRE, all metrics provide a lower score for higher readability, while the human judgments provide a higher score for higher readability. To calculate correlations, we multiply the scores by $-1$ (except for LW and FRE), so that text rated as more readable by traditional metrics will be positively correlated with human judgments.

\subsection{LMs as evaluators of readability \hyperref[rq3]{\rqthree}}\label{sec:rq-three-setup}

We experiment with the following 5 LMs as evaluators of readability: Mistral 7B~\cite{Jiang2023Mistral7}, Mixtral 7B~\cite{jiang2024mixtral}, Gemma 7B~\cite{gemma_2024}, Llama 3.1 8B, and Llama 3.3 70B~\cite{dubey2024llama3herdmodels}. We experiment with 3 prompts and report the prompts in \cref{appendix:prompt}. We report the Pearson and Kendall-Tau correlations of the scores provided by each LM with the human judgments.

\subsection{Analysis of summarization datasets \hyperref[rq4]{\rqfour}}
% \daniel{Consider adding paragraph heading to the next two paragraphs so that it's clear what each is about.}
To test the ability of our results to generalize to datasets outside of the one collected by \citet{August2024KnowYA}, we include datasets with intended audiences more specific than ``general'' - experts and kids. We expect expert-targeted datasets to be given low readability scores and kid-targeted datasets to have high readability scores.

\paragraph{Expert targeted datasets.} We include 3 expert-targeted datasets: arXiv, PubMed~\cite{Cohan2018ADA} and SciTLDR~\cite{cachola-etal-2020-tldr}.
arXiv and PubMed are collections of abstracts in the Computer Science and Biomedical domains, respectively~\cite{Cohan2018ADA}. SciTLDR is a collection of short, expert-targeted, one sentence summaries of Computer Science papers. We expect our methods to provide low readability scores. Additionally, the comparison of SciTLDR to arXiv and PubMed allows us to test if the scores are length dependent. 

\paragraph{Kid-targeted dataset.} The Science Journal for Kids (SJK) dataset is a collection of summaries of scientific papers in a variety of domains, intended for kids~\cite{BioLaySumm_2024}. 
% We noticed the original SJK dataset published by~\citet{BioLaySumm_2024} contained a number of grammatical mistakes, likely a result of PDF processing errors. We recollected the dataset by accessing the summaries directly on the SJK website\footnote{\url{https://www.sciencejournalforkids.org/}} to ensure that the summaries are grammatically correct. 
Given that this dataset is targeted to kids, we expect it would receive high readability scores.

\paragraph{General audience datasets.} In addition to the datasets listed above, we evaluate 6 popular datasets intended for PLS: CDSR~\cite{Guo2020AutomatedLL}, PLOS~\cite{Goldsack2022MakingSS}, eLife~\cite{Goldsack2022MakingSS}, Eureka~\cite{Zaman2020HTSSAN},  CELLS~\cite{Guo2022RetrievalAO}, and SciNews~\cite{pu-etal-2024-scinews}. These datasets are intended for a general audience. CDSR, PLOS, and CELLS are written by journal editors or experts. eLife Sciences gives paper authors the option to write ``eLife digests,'' with the goal of ``cutting jargon and putting research in context.''
\footnote{\url{https://elifesciences.org/digests}} 
The Eureka dataset was collected from EurekaAlert, which hosts press releases about research for scientific journalists.
% \footnote{\url{https://www.eurekalert.org/}} 
Finally, SciNews is a collection of scientific news reports, written by science reporters.

\Cref{tab:dataset-comp} contains a comparison of the summarization datasets analyzed in this paper. We use the test split of each dataset for our analysis and we report the intended audience, domain, number of documents in the test set, and average number of white-space delineated tokens.
In total, we analyze 10 popular scientific summarization datasets.

\input{tables/dataset-comparison}

%% file: tables/dataset-comparison.tex
\begin{table}[t]
\setlength{\tabcolsep}{3pt}
\centering \footnotesize
\begin{tabular}{r|lll l}
\textbf{Dataset} & \textbf{Audience} & \textbf{Domain} & \textbf{\# Docs} & \textbf{\# Tokens} \\ \hline
% \multicolumn{4}{r}{\textit{Readability Known Datasets}} \\ \hdashline[0.5pt/3pt]
arXiv & Experts & CS &6440 & 163 \\
PubMed & Experts & Medicine &6658 & 205 \\
SciTLDR & Experts & CS &618 & 19 \\
SJK & Kids & Varied & 284  & 142 \\ \hdashline[0.5pt/3pt]
% \multicolumn{4}{r}{\textit{PLS Datasets}} \\ \hdashline[0.5pt/3pt]
CDSR & General & Healthcare &284 & 221 \\
PLOS & General & Biomed &1376 & 195 \\
eLife & General & Biomed &241 & 383 \\
Eureka & Journalists & Varied &1010 & 662 \\
CELLS & General & Biomed &6311 & 162 \\
SciNews & General &  Varied & 4188 & 615 
\end{tabular}
\caption{Comparison of the datasets analyzed in this paper. The first 4 are datasets in with a specific target audience. The following 6 datasets are commonly used in PLS literature. We report the number of documents (\# Docs) in the test set as well as the average number of tokens (\# Tokens).
% \daniel{FYI: change D and T to Docs/Tokens since you haves room.}
}\label{tab:dataset-comp}
\vspace{-4mm}
\end{table}

%% file: 4-Results.tex
%Below is a summary of the results from the experiments described in \S\ref{sec:experimental-setup}. 

\begin{figure}[t]
    \centering
    \includegraphics[width=0.99\linewidth]{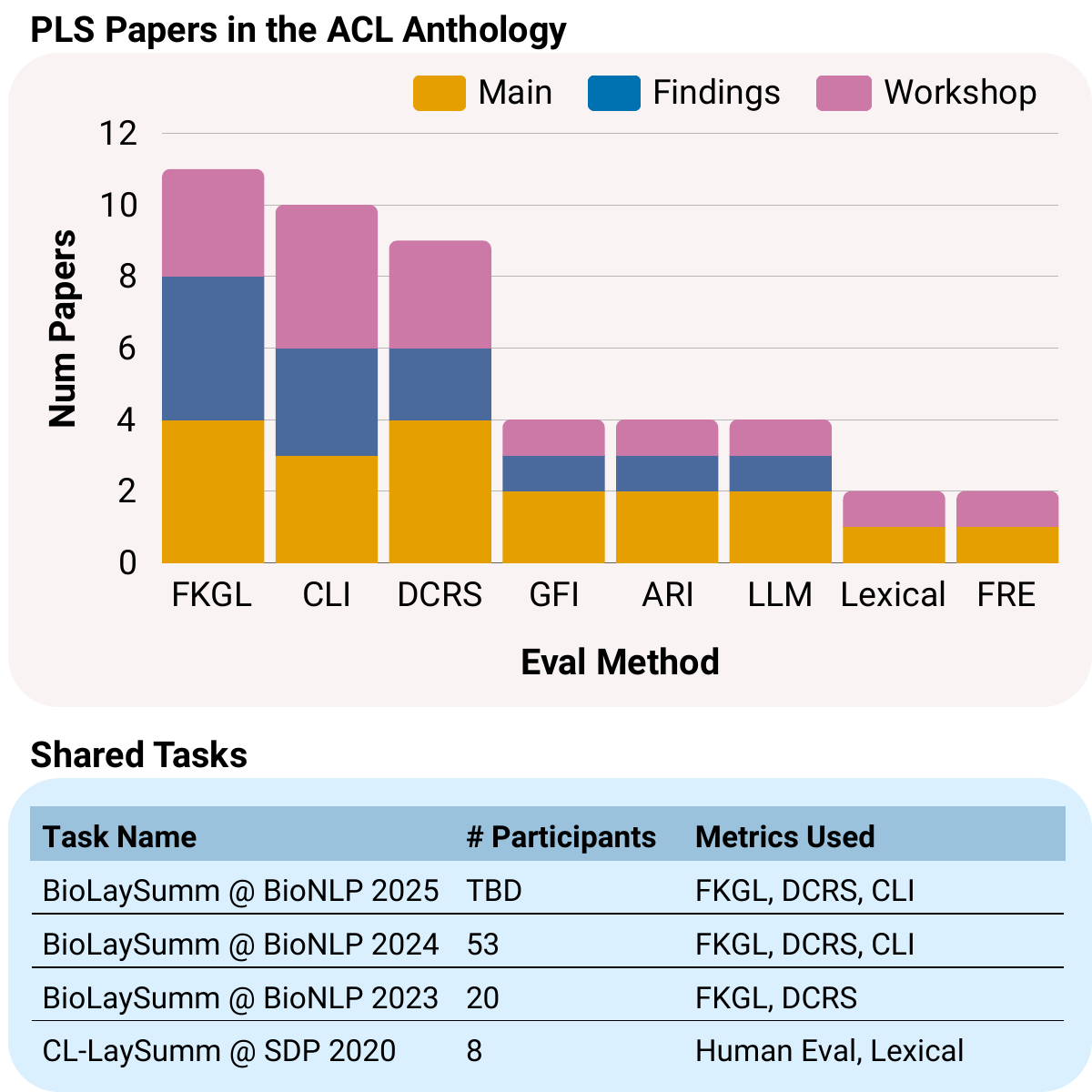}
    \vspace{-5mm}
    \caption{Evaluation metrics used by papers published in the ACL Anthology. We report the count of papers using each method out of a total of 18 papers. We additionally report the evaluation strategies used by PLS shared tasks and the number of participants.}
    \label{fig:lit-survey-results}
    \vspace{-4mm}
\end{figure}
\section{Results}\label{sec:results}

\subsection{Current PLS evaluation standards \hyperref[rq1]{\rqone}}\label{subsec:results-rqone}
We found 18 ACL Anthology papers on the task of PLS and 3 shared tasks, representing 81 additional papers. \Cref{fig:lit-survey-results} shows the literature survey results, excluding metrics used by a single paper. FKGL is the most popular metric, followed by CLI and DCRS. LM-based evaluations are uncommon (4 of the 18 papers). The shared task BioLaySumm used FKGL and DCRS for both years, adding CLI in 2024. BioLaySumm 2025 is ongoing at the time of writing; the organizers plan to use FKGL, DCRS, and CLI. Our survey shows that PLS is an increasingly popular topic of study, as the number of participants in shared tasks increased from 8 in 2020 to 53 in 2024, emphasizing the importance of PLS evaluation. Less popular metrics include GFI, ARI, lexical proxies (e.g., number of sentences in a document), and FRE. In \cref{subsec:results-rqtwo}, we place the highest importance on the results of the most commonly used evaluation metrics: FKGL, CLI, and DCRS.

\subsection{Comparing traditional readability metrics to human judgments \hyperref[rq2]{\rqtwo}}\label{subsec:results-rqtwo}

In \Cref{subtab:trad-corr-results}, we report the Pearson and Kendall-Tau correlation of 8 traditional readability metrics with human judgments. We find that 6 of the 8 metrics have less than $0.3$ Pearson correlation with human judgments. DCRS and CLI have the highest correlation, achieving $0.2$ Pearson points higher correlation than the most popular metric, FKGL (\Cref{subsec:results-rqone}). FKGL receives only $0.16$ Pearson and $0.08$ Kendall-Tau correlation, indicating little to no correlation with human judgment. 

\Cref{tab:example-summ} shows an example summary and readability scores, along with its human judgment. The human annotators gave the example summary an average rating of 4.05/5; they found the text fairly readable. However, the majority of traditional metrics give the summary poor readability scores: college level or higher. This is likely because the text includes domain-specific vocabulary, such as ``acute respiratory distress syndrome (ARDS),'' which is penalized by traditional metrics. 
Traditional readability metrics do not account for elements of the summary that make it more readable, such as defining ARDS as ``a very serious lung disease'' and explaining the scientists' motivation to ``test a new method of lung damage diagnosis.''

\input{tables/human_correlation}

\input{tables/summ-examples}
\begin{figure}[t]
    \centering
    \vspace{-2mm}
    \includegraphics[width=.99\columnwidth]{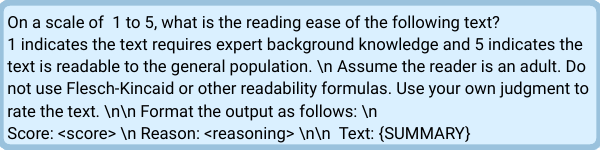}
        \caption{The best performing prompt of the 3 we tested. We report the results of this prompt in~\Cref{subtab:lm-corr-results} and the results of the remaining prompts in~\Cref{appendix:prompt}.}
    \label{fig:best-prompt}
    \vspace{-4mm}
\end{figure}

\subsection{LMs as evaluators of readability \hyperref[rq3]{\rqthree}}\label{subsec:results-rqthree}
Traditional readability metrics rely on lexical proxies and do not measure other elements of a summary that could make it more readable, such as definitions of technical terms, explanations of important concepts, or descriptions of impact and motivation. LMs have been shown to perform well on many language understanding tasks~\cite{Brown2020LanguageMA,srivastava2023beyond}, indicating that they have some understanding of language. We hypothesize that this knowledge will translate well to the task of PLS, and the LMs will be able to reason about more complex features of a summary that impact the readability.

We experiment with 3 prompts. We report best performing prompt in~\Cref{fig:best-prompt} and its results in~\Cref{subtab:lm-corr-results}; the other prompts and their results are in~\Cref{appendix:prompt}.  All of the LMs outperform the traditional metrics in correlation with human judgments. The best performing model, \texttt{Llama 3.3 70B}, outperforms the best traditional metric, DCRS, by nearly $0.2$ Pearson points. We conduct significance testing and report the p-values comparing the LM results to the traditional metrics in \Cref{appendix:sig-testing}. 

Performance in this task is not solely a factor of model size, as we see that smaller models perform similarly to the larger models. The difference in performance between the LMs is small, indicating that most generally well-performing models can be good judges of readability.

\Cref{tab:example-summ} contains an example summary and its associated scores from each LM. The human annotators rated the example summary a 4.05 out of 5 on reading ease. All models gave the summary a rating of 4 or 4.5  out of 5. The reasoning provided by \texttt{Llama 3.3 70B} states that the ``concepts discussed, such as analyzing breath samples and identifying chemical compounds, are also explained in a way that is easy to understand.''
The model notes that the summary ``may require some effort and attention,'' contributing to the model's reasoning for assigning the summary a 4/5 rather than a 5/5. This output indicates that the model is using its language reasoning abilities to rate the summary on attributes deeper than lexical features.

\begin{figure*}[!t]
\centering\includegraphics[width=.99\textwidth]{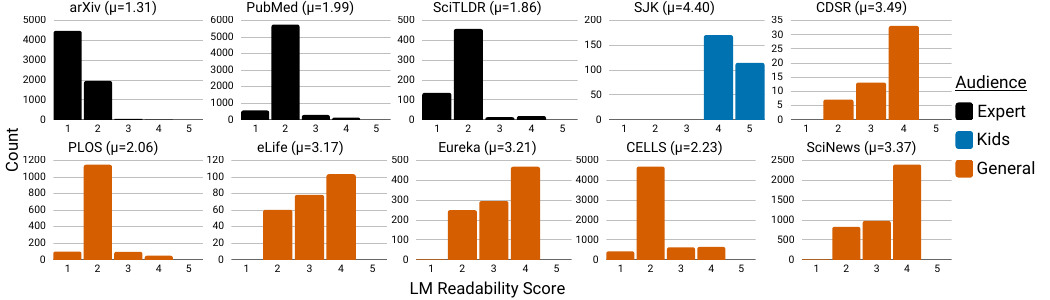}
    \vspace{-3mm}
    \caption{Histogram of LM readability scores and the mean scores ($\mu$) for each dataset, as judged by \texttt{Llama 3.3 70B}. As we can see from the results, PLOS and CELLS are judged to be similarly readable to the expert targets datasets (arXiv, PubMed, and SciTLDR). The most readable PLS datasets are CDSR and SciNews.}
    \vspace{-4mm}
    \label{fig:histogram-datasets}
\end{figure*}

\input{tables/lm-readability-scores-by-dataset}
\subsection{Analysis of summarization datasets \hyperref[rq4]{\rqfour}}\label{subsec:results-rqfour}
We analyze scientific summarization datasets using the LM evaluators. We use \texttt{Llama 3.3 70B}, the best performing model from \Cref{subsec:results-rqtwo}. In \Cref{fig:histogram-datasets}, we include histograms of the readability scores for all 10 tested datasets, to visualize the distributions. 
In~\Cref{tab:lm-readability-scores-by-dataset}, we report the mean, median, and variance of the readability scores for each dataset.

% In \Cref{appendix:additional-stats}, we report the mean, median, and variance of the LM readability scores. 

We've shown that most LM judgments of readability correlate higher with human judgments than traditional metrics. In order to further validate our findings, we begin our analysis with 4 datasets with specific target audiences - experts or kids. 

\paragraph{Expert-targeted datasets.} We experiment with 3 datasets intended for expert readers: arXiv, PubMed, and SciTLDR. ArXiv, PubMed, and SciTLDR receive low readability scores, averaging less than 2/5. This matches our expectations since summaries intended for an expert audience typically have low readability for non-experts. We also note that SciTLDR receives similarly low readability scores, despite containing significantly shorter summaries than the arXiv and PubMed datasets. This shows that the LM evaluator is not simply favoring shorter summaries as more readable. 

\paragraph{Kid-targeted dataset.} SJK receives high readability scores, with an average readability of $4.40$. The results of the expert and kid targeted datasets match our expectations of readability scores, and serve to support the analysis of the remaining 6 general-audience datasets below.

\paragraph{General audience datasets.} We analyze 6 popular PLS datasets: CDSR, PLOS, eLife, Eureka, CELLS, and SciNews. PLOS and CELLS receive mean readability scores of 2.06 and 2.23, respectively. These scores are similar to the expert-targeted datasets described above, indicating that these two datasets may not be well-suited for PLS. SciNews and CDSR receive the highest readability scores, with average scores of 3.49 and 3.37, respectively, indicating that they are the well suited for the task of PLS.

\begin{figure}[!h]
    \centering
    % First subfigure
    \begin{subfigure}[b]{0.42\textwidth}
        \centering
        \includegraphics[width=\textwidth]{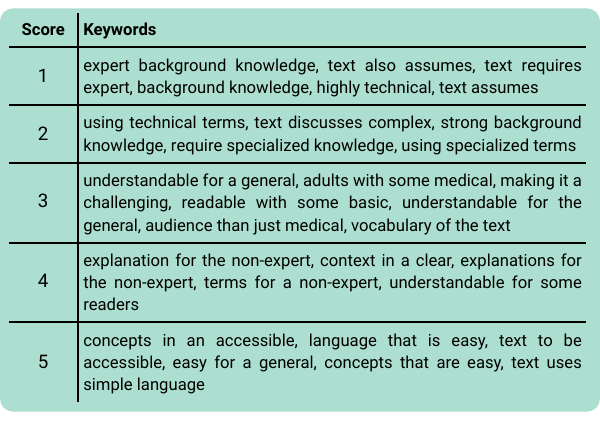}
        \vspace{-6mm}
        \caption{Keywords stratified by score.}
        \label{fig:keywords-score}
    \end{subfigure}
    \hfill
    % Second subfigure
    \begin{subfigure}[b]{0.42\textwidth}
        \centering
        \includegraphics[width=\textwidth]{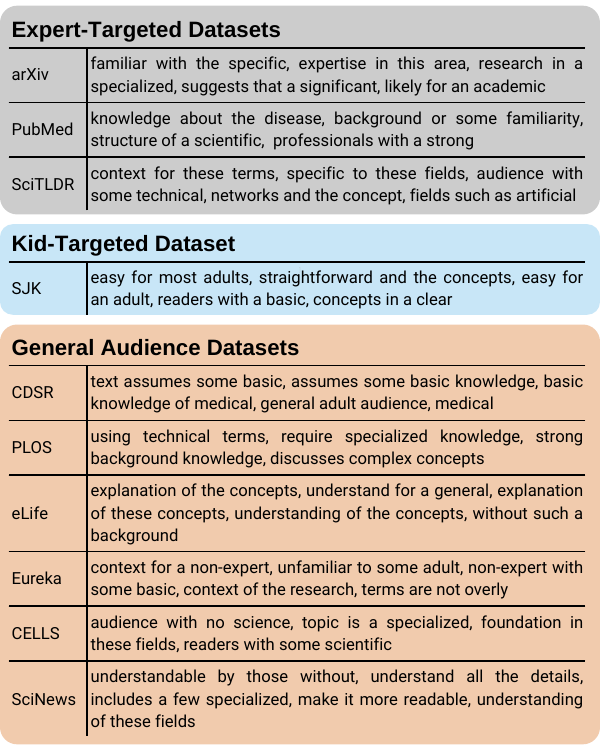}
        \vspace{-6mm}
        \caption{Keywords stratified by dataset.}
        \label{fig:keywords-dataset}
    \end{subfigure}

    \caption{Keywords mentioned in the reasoning of the LM evaluator for why a summary was given a certain readability score. \Cref{fig:keywords-score} contains the keywords stratified by score and \Cref{fig:keywords-dataset} contains keywords stratified by dataset.}
    \label{fig:keyword-analysis}
    \vspace{-6mm}
\end{figure}

\paragraph{Keyword analysis.} To understand the LM's reasoning for assigning scores, we use the \texttt{YAKE!} algorithm to extract keywords from the reasoning provided by the LM evaluator for why each summary was provided with a specific score~\cite{CAMPOS2020257}. \Cref{fig:keywords-score} contains the keywords stratified by score and \Cref{fig:keywords-dataset} contains the keywords stratified by dataset. 
When stratified by score, the model mentions issues such as requiring ``expert background knowledge'' and ``using specialized terms'' for summaries with readability scores of 1 or 2. For summaries with scores of 4 or 5, the model references how the summaries include ``explanations for the non-expert'' and explains ``concepts in an accessible'' manner.
When stratified by dataset, for datasets with generally low readability scores, such the model mentions issues such as requiring ``specialized knowledge'' or that the text is ``likely for an academic.'' The model also mentions the domain specific knowledge required such as Pubmed's focus on ``disease[s].'' For datasets with generally high readability scores, such as SJK and SciNews, the model mentions how the summaries are ``easy for most adults'' and how the text is ``understandable by those without'' background knowledge. This keyword analysis indicates LMs are attributing their judgements to deeper attributes that contribute to readability compared to traditional metrics.  

% We report the extracted keywords and conduct a similar analysis, but with the keyword extraction stratified by dataset, in \cref{appendix:keyword-analysis-dataset}. LMs are attributing their judgements to deeper attributes that contribute to readability compared to traditional metrics. 

\paragraph{LM evaluators vs. traditional metrics.} Finally, we compare the results of the analysis using traditional metrics to LM evaluators of readability. In this analysis, we focus on \texttt{Llama 3.3 70B}, the best performing LM, and FKGL, the most popular readability metric. \Cref{tab:ranked-datasets} compares the average LM readability and FKGL score for each dataset, and how each metric would rank the datasets. All but 1 dataset changed their ranking depending on the metric used. arXiv has the largest delta, ranking 10th in readability according to the LM evaluator and 2nd according to FKGL. FKGL ranking arXiv as the 2nd most readable is particularly concerning, as this dataset is a collection of scientific abstracts, intended for an expert audience.
To measure disagreement, we convert each metric into binary scores of ``high readability'' and ``low readability.'' For FKGL, we consider any summary given a score of under 12 points to have high readability. FKGL considers any text above 12 to be college reading level. For the LM evaluator, we consider any summary given a score of 3 or higher to have high readability. By converting the scores to binary labels, we calculate the Cohen's Kappa score~\cite{McHugh2012InterraterRT} for agreement as 0.17, indicating the two metrics have fair but not substantial agreement. We provide examples of this disagreement in \Cref{tab:disagreement-example}. This analysis shows how the evaluation metrics we use can greatly influence the conclusions we draw. 

\input{tables/ranked-datasets}

\input{tables/disagreement-example}

%% file: tables/human_correlation.tex
\begin{table}[!t]
\centering
    \begin{subtable}{.49\textwidth}
        \centering \small
        \begin{tabular}{l|ll}
            \textbf{Metric} & \textbf{Pearson} & \textbf{Kendall Tau} \\ \hline
             FKGL &  0.16 & 0.08 \\
            CLI & 0.36 & 0.20 \\
            \textbf{DCRS} & \textbf{0.37} & \textbf{0.24} \\
            GFI & 0.21 & 0.11 \\
            ARI & 0.10 & 0.02 \\
            FRE & 0.29 & 0.15 \\
            Spache & 0.13 & 0.04 \\
            LW & -0.06 & -0.03 \\
        \end{tabular}
        \vspace{-1mm}
        \caption{Traditional metric scores correlation with human judgment.}
        \label{subtab:trad-corr-results}
        \vspace{2mm}
    \end{subtable}    
    % \hspace{1mm}
    \begin{subtable}{.49\textwidth}
        \centering \small
            \begin{tabular}{l|ll}
            \textbf{Model} & \textbf{Pearson} & \textbf{Kendall Tau} \\ \hline
            % OLMo 2 7B & 0.15 & 0.10 \\
            Mistral 7B & 0.52 & 0.44 \\
            Mixtral 7B & 0.54 & 0.41 \\
            Gemma 7B & 0.54 & 0.43 \\
            Llama 3.1 8B & 0.45 & 0.34 \\
            \textbf{Llama 3.3 70B} & \textbf{0.56} & \textbf{0.35} \\
            % \textbf{OLMo 2 13B} & \textbf{0.57} & \textbf{0.47} \\
        \end{tabular}
        % \vspace{2.3mm}
        \vspace{-1mm}
        \caption{LM scores correlation with human judgment.}\label{subtab:lm-corr-results}
    \end{subtable} 
    % \vspace{-7mm}
    \vspace{-6mm}
    \caption{We report the Pearson and Kendall-Tau correlation of each metric with human judgment. Tab.\ref{subtab:trad-corr-results} contains the correlation of traditional readability metrics with human judgment. DCRS and CLI have the highest correlation with human judgment. Notably, the most popular metric, FKGL, as shown in \S\ref{subsec:results-rqone}, has low correlation with human judgment. Tab.\ref{subtab:lm-corr-results} contains the correlation of LM models as evaluators with human judgment. All 5 models achieve higher correlation than all of the traditional metrics.
    }
    % \vspace{-4mm}
    \label{tab:corr-results}

\end{table}

%% file: tables/summ-examples.tex
\begin{table*}[!htb]
\centering
    \begin{subtable}{.95\textwidth}
      \centering 
        \begin{tabular}{p{\textwidth}}
            \textexample{Scientists create a device which can detect the onset of acute respiratory distress syndrome (ARDS), a very serious lung disease, by measuring chemicals in patients’ exhaled breath The researchers wanted to test a new method of lung damage diagnosis by analyzing patient breath samples. In particular, the researchers were looking for better ways to detect acute respiratory distress syndrome (ARDS), a form of lung injury that causes inflammation and severe damage. [...]  a much larger group of test subjects is necessary to further validate their method. This new method of breath analysis could be a noninvasive, cost effective way to diagnose and track ARDS, and could potentially be modified to screen for other serious conditions as well.}
        \end{tabular}
        \vspace{-4mm}
        \caption{Excerpt of an example summary. This summary is written by an expert and is labeled as a low complexity summary.}\label{subtab:ex-summ}
    \end{subtable}%
    \vspace{1mm}
    \quad
    \begin{subtable}[t]{.65\textwidth}
        \centering \footnotesize
        \begin{tabular}{r|lll}
        Metric & Score & S-12 & US Grade Level \\ \hline
        FKGL $\downarrow$ & 13.9 & 12 & College \\
        CLI $\downarrow$ & 12.7 & 12 & College \\
        DCRS $\downarrow$ & 11.3 & 8.9 & College graduate \\
        GFI $\downarrow$ & 18.6 & 12 & Above college graduate \\
        ARI $\downarrow$ & 16.7 & 13 & College graduate \\
        FRE $\uparrow$ & 50.2 & 50 & 12th grade \\
        Spache $\downarrow$ & 8.7 & 12 & 9th grade \\
        LW $\uparrow$ & 19.5 & 60 & College graduate \\

        \end{tabular}
        \vspace{-1mm}
        \caption{Scores given be each metric for the example summary. $\downarrow$ indicates a lower score is more readable while $\uparrow$ indicates a higher score is more readable. We provide ``S-12'', the score each metric would assign US grade 12, to help contextualize the scores. We additionally translate each score to the US grade level.}
        \label{subtab:ex-metric-scores}
    \end{subtable} 
    \quad
    \begin{subtable}[t]{.32\textwidth}
     \centering \footnotesize
        \begin{tabular}{r|l}
        \multicolumn{2}{c}{} \\
        \multicolumn{2}{c}{} \\
        \multicolumn{2}{c}{} \\
        Model & Score \\ \hline
        Mistral 7B & 4 \\
        Mixtral 7B & 4.5 \\
        Gemma 7B & 4 \\
        Llama 3.1 8B & 4 \\
        Llama 3.3 70B & 4 \\
        \end{tabular}
    \vspace{-1mm}
    \caption{Scores given be each model for the example summary %\daniel{from X dataset}. 
    The scores are on a scale of 1-5, with 5 being the most readable.}
    \label{subtab:ex-lm-scores}
    \end{subtable} 
    \vspace{-2mm}
    \caption{ \ref{subtab:ex-metric-scores} contains an example summary from~\citet{August2024KnowYA}'s dataset. \ref{subtab:ex-metric-scores} contains each metric's score for the example summary. \ref{subtab:ex-lm-scores} contains each model's readability scoring for the example summary. On average, the human annotators rated this summary a 4.05/5, indicating they found the summary fairly readable. All the LM evaluators rate the summary a 4 or 4.5 out of 5, agreeing with the human annotators. In contrast, 6 out of 8 of the traditional metrics rate the summary at a college reading level or higher, which is considered low readability. 
    % \daniel{Say how this is different from metrics/LLMs judgement?}
    }
    \label{tab:example-summ}
    \vspace{-4mm}
\end{table*}

%% file: tables/lm-readability-scores-by-dataset.tex
\begin{table}[t]
\centering \footnotesize
\begin{tabular}{r|ccc}
Dataset & Mean & Median & Var \\ \hline
arXiv & 1.31 & 1 & 0.23 \\
PubMed & 1.99 & 2 & 0.19 \\
SciTLDR & 1.86 & 2 & 0.32 \\
SKJ & 4.40 & 4 & 0.24 \\ \hdash
 CDSR & 3.49 & 4 & 0.52 \\
PLOS & 2.06 & 2 & 0.26 \\
eLife & 3.18 & 3 & 0.65 \\
Eureka & 3.21 & 3 & 0.67 \\
CELLS & 2.23 & 2 & 0.50 \\
SciNews & 3.37 & 4 & 0.64
\end{tabular}
\vspace{-2mm}
\caption{Readability scores on a scale of 1 to 5, as judged by \texttt{Llama-3.3-70B}, 5 being the most readable. We report the mean, median, and variance of each score.}\label{tab:lm-readability-scores-by-dataset}
\vspace{-4mm}
\end{table}

%% file: tables/ranked-datasets.tex
\begin{table}[t!]
\vspace{-2mm}
\centering \small
\begin{tabular}{r|ll|lll}
 & \multicolumn{2}{c|}{LM Eval} & \multicolumn{3}{c}{FKGL} \\
Dataset & S & R & S & R & $\Delta$R \\ \hline
arXiv & 1.31 & 10 & 11.53 & 2 & $+8$ \\
PubMed & 1.99 & 8 & 14.14 & 5 & $+3$ \\
SciTLDR & 1.86 & 9 & 15.66 & 10 & $-1$ \\
SKJ & 4.40 & 1 & 8.41 & 1 & $0$ \\
CDSR & 3.49 & 2 & 14.08 & 4 & $-2$ \\
PLOS & 2.06 & 7 & 15.44 & 9 & $-2$ \\
eLife & 3.18 & 5 & 11.87 & 3 & $+2$ \\
Eureka & 3.21 & 4 & 14.87 & 6 & $-2$ \\
CELLS & 2.23 & 6 & 15.35 & 8 & $-2$ \\
SciNews & 3.37 & 3 & 14.98 & 7 & $-4$
\end{tabular}
\vspace{-2mm}
\caption{The mean score (S) and rank (R) for each dataset, as judged by an LM evaluator and FKGL. $\Delta$R represents the change in rank from the LM evaluator to the FKGL scores.}\label{tab:ranked-datasets}
\vspace{-4mm}
\end{table}

%% file: tables/disagreement-example.tex
\begin{table*}[!t]
\centering
    \begin{subtable}{.46\textwidth}
        \centering \footnotesize
        \begin{tabular}{p{.99\textwidth}}
        \textexample{Wind power is an important source of renewable energy, but some people are concerned that conventional wind turbines are too loud and too hazardous for birds and bats. We wanted to create a new kind of wind energy harvesting machine based on the jiggling motion of cottonwood tree leaves in the wind, which would be quieter and safer for wildlife. After building and testing artificial cottonwood leaves that moved and created electricity in the wind, we found that they didn’t produce enough energy to feasibly use for electricity production. We also tried building a cattail-like device to generate electricity when it swayed in the wind, [...]
        % but it also didn’t produce enough energy to make it reasonable to use. 
        % Though our research showed that artificial plants’ jiggling or swaying isn’t likely to be a cost-effective way to produce electricity, we think it could be fruitful to look into other plant-inspired designs for harvesting wind energy. We also are testing a previously unexploited biological material known to convert mechanical to electrical energy far more effectively than the ones used today.
        } 
        \end{tabular}
        \vspace{-3mm}
        \caption{FKGL = 16.47 (College-graduate), LM score = 4/5.}
        \label{subtab:ex-disaggreement-sjk}
    \end{subtable}    
    % \hspace{1mm}
    \begin{subtable}{.51\textwidth}
        \centering \footnotesize
        \begin{tabular}{p{.99\textwidth}}
        \textexample{Introduction. Accumulation of glycochenodeoxycholic acid (GCDC) in serum has a clinical significance as an inductor of pathological hepatocyte apoptosis, which impairs liver function. Inhibition of GCDC accumulation can be used as a marker in therapy. This study was aimed to quantify the serum level of GCDC in obstructive jaundice patients. Methodology. GCDC acid level in the serum was quantified using high performance liquid chromatography (HPLC) technique according to Muraca and Ghoos modified method. It was performed before and after decompression at day 7 and day 14. The sample was extracted with solid phase extraction (SPE) technique on SPE column. The results were analyzed using SPSS V 16.0 (P < 0.05) [...]
        % and quantified with standard curve on GCDC acid. 
        % Result. There were 21 cases with range of GCDC acid serum level before decompression was 90.9 (SD 205.5) μmol/L and day 7 after decompression decreased to 4.0 (SD 46.4) μmol/L and then increased to 11.3 (SD 21.9) μmol/L (P < 0.05).
        % This method could separate GCDC acid on serum with good resolution, high precision and accuracy, and linear calibration curve on measured level range. Conclusion. HPLC can quantify GCDC acid serum on obstructive jaundice patients and can be used to support its pharmacokinetic study.
        }
        \end{tabular}
        \vspace{-3mm}
        \caption{FKGL = 10.0 (10th grade), LM score = 1/5.}
        \label{subtab:ex-disaggreement-plos}
    \end{subtable} 
    % \vspace{-2mm}
    \caption{Examples of disagreement between FKGL and the LM evaluator. \ref{subtab:ex-disaggreement-sjk} contains an example from the SJK dataset that the LM rated high readability and FKGL rated low readability. \ref{subtab:ex-disaggreement-plos} contains a summary from the Pubmed dataset that the LM rated low readability while FKGL rated high readability.}\vspace{-4mm}
    \label{tab:disagreement-example}
\end{table*}

%% file: 5-Discussion.tex
\section{Discussion}\label{sec:discussion}

We found PLS an increasingly popular area of study, but researchers primarily rely on a handful of traditional metrics for evaluation. However, we found that traditional metrics are imperfect measures of readability and LM evaluators can draw significantly different, and more accurate, conclusions about PLS datasets than when using FKGL, the most common metric.

\subsection{Why traditional readability metrics are insufficient measures of readability}\label{subsec:discussion-inconsistency}
% \daniel{if this does not need to be a "subsection" consider turning it into a paragraph and it will save you some space. }
We consider 2 explanations for the poor correlation of readability metrics with human judgments: definitional inconsistency or measurement error. Definitional inconsistency means that the definition of ``readable,'' as measured by the metrics, differs from the definition of ``readable,'' as considered by human judges. Measurement error means that, even if we have the correct definition, we are not measuring readability properly.
%This distinction is important because it allows us to understand the source of the low performance and design appropriate solutions.
We argue that there is evidence for both problems.

On definitional inconsistency, the majority of readability metrics originated in the education domain. Traditional readability metrics typically define a ``readable'' text as one with an appropriate text complexity for the number of years of education (i.e., a text has a US 9th grade reading level)~\citep{Gunning1968TheTO,Coleman1975ACR,flesch1952simplification}. In contrast, the field of PLS typically defines a ``readable'' text as one that gives a non-expert, adult reader an overall understanding of the source article. These different definitions have different implications for the resulting text. If optimizing for education-appropriate text complexity, we can measure the complexity of the vocabulary or sentences. However, using the PLS definition of readability, we should measure features such as whether the text includes explanations of technical terms or how much background is required to understand the concepts.

Traditional readability metrics also suffer from measurement error. Even if we assume a consistent definition, traditional metrics do not properly measure readability. They measure lexical properties, such as number of syllables in a word, which penalizes summaries for using clearly defined technical terms. Traditional metrics also do not measure deeper features that contribute to readability, such as how much background is required to understand the text. In \Cref{subsec:results-rqthree}, we show that LMs are better able to reason over these more complex attributes. 

\Cref{tab:disagreement-example} shows examples in which the LM evaluator and FKGL disagree on the readability. FKGL rates a summary from the SJK dataset as having a graduate-college reading level, while the LM rates it as highly readable (\Cref{subtab:ex-disaggreement-sjk}). Although the summary explains the concepts well, long words such as ``harvesting'' and ``electricity'' likely cause FKGL to rate the summary as less readable. \Cref{subtab:ex-disaggreement-plos} has a Pubmed example, which the LM rates as having low readability, while FKGL assigns the summary a 10th grade reading level. This example contains many short words, such as ``GCDC'' and ``SPE'', which are favored by FKGL. Although short, these technical words that are not well defined. For example, the  ``GCDC'' is defined as ``glycochenodeoxycholic acid,'' but  is not otherwise explained. In general, we  notice that FKGL favors acronyms, which are often present in technical text.

\subsection{Recommendations and Future Directions}\label{subsec:recommendations}
We find that many traditional readability metrics have poor correlation with human judgments and that LMs provide better judgments. However, LM-evaluators are an imperfect solution since they are subject to bias and a lack of interpretability~\cite{Liu2023LLMsAN, Wang2023LargeLM, Shen2023LargeLM, Stureborg2024LargeLM}. Therefore, we recommend a multi-faceted evaluation of PLS that uses a combination of traditional readability metrics and LM evaluators. Specifically, we recommend using DCRS and CLI, which have the highest correlation with human judgments. We recommend discontinuing use of FKGL for PLS, the current most popular metric, due to low correlation with human judgment. We recommend using LMs as additional metrics, especially for more qualitative evaluations, such as the keyword analysis conducted in \Cref{subsec:results-rqthree}. These types of analyses give a more holistic view of the benefits and downsides of datasets and methods. Finally, we recommend that PLS research use datasets with higher readability scores (\Cref{subsec:results-rqthree}), such as CDSR and SciNews. We recommend that PLOS and CELLS be considered general scientific summarization datasets and not plain language datasets. This recommendation is particularly impactful as PLOS has been used in every year the shared task BioLaySumm has occurred~\cite{goldsack-etal-2024-biolaysumm,biolaysumm-2023-overview}.

Future work should focus on constructing metrics that better align with human judgments of readability in both definition and measurement (\Cref{subsec:discussion-inconsistency}). We show that LMs are promising and worthy of future work that can decrease bias and improve interpretability. Dataset collection should focus on collecting highly readable summaries and consider deeper attributes of readable summaries, such as explanations of technical concepts.

%% file: appendix.tex
\appendix

\begin{center}
% \onecolumn
{\Large \textbf{Appendix}}
% \twocolumn
\end{center}

\section{Human annotated dataset details}\label{appendix:dataset_details}
We use the human annotated data collected by~\citet{August2024KnowYA}. The dataset includes 60 summaries over 10 papers, 6 summaries per paper. Of the 6 summaries, 2 are written by experts and 4 are machine written by GPT3. The 10 papers were sampled from the top 10\% of papers from \textit{r/science}, a subreddit dedicated to public discussions of scientific papers. These papers were chosen as a proxy for scientific topics the general public is most interested in. The dataset was annotated by 593 Mechanical Turk workers in total across the three tasks in the original study.
\Cref{tab:human_annotation_distribution} contains the distributions of scores assigned by the human annotators.
\vspace{-5mm}
\input{tables/human_annotation_distribution}

In order to measure inter-annotator agreement, we bin the scores into a binary ``high-readability'' and ``low readability.'' Summaries given scores of 3 or higher are considered highly readable while summaries assigned scores less than 3 are considered to have low readability. We use Cohen's Kappa to calculate an inter-annotator agreement of 0.6. This is a moderate agreement for a somewhat subjective task, indicating that there is some general notion of readability. We also note that this is significantly higher than the agreement between traditional metrics and LMs (0.17 as shown in~\Cref{subsec:results-rqfour}).

% \section{Summarization dataset comparison}\label{appendix:dataset_comparison}
% \Cref{tab:dataset-comp} contains a comparison of the summarization datasets analyzed in this paper. We use the test split of each dataset for our analysis. For each dataset, we report the intended audience, domain, number of documents in the test set, and average number of white-space delineated tokens.
% \input{tables/dataset-comparison}

\section{LM readability evaluation prompts}\label{appendix:prompt}
We experiment with 3 prompts, shown in~\Cref{tab:prompts}. The \texttt{Simple Prompt} simply asks the LM to rate the text for reading ease on a scale of 1 to 5. The American Society for Cell Biology (ASCB) provides guidelines for best practices in scientific communication.\footnote{\href{https://www.ascb.org/science-policy-public-outreach/science-outreach/communication-toolkits/best-practices-in-effective-science-communication/}{ASCB Best Practices in Science Communication
}} In the  \texttt{ASCB Prompt}, we provide these guidelines to the LM as guidance for rating the readability. Finally, the  \texttt{Own Reasoning Prompt} is similar to the  \texttt{Simple Prompt}, but with the additional instruction for the LM to use it's own judgment to rate the text, rather than traditional readability formulas, such as FKGL. 

\input{tables/prompt-comparison}

We report the Pearson and Kendall-Tau correlation of each prompt with human judgment in \Cref{tab:prompt-comparison}. The \texttt{Own Reasoning Prompt} performs the best when averaged across all models. We found that the models tended to over-rely on the guidance provided in the \texttt{ASCB Prompt}, providing lower scores if the conditions are not met. For the \texttt{Simple Prompt}, the models would occasionally try to calculate FKGL or another readability metric, rather than using its own reasoning. This is likely because FKGL is strongly associated with readability in the models' training data. We found that the \texttt{Own Reasoning Prompt} struck the right balance between providing enough instructions that the model is able to understand the task without providing too much information for the model to over-rely on. However, it is notable that the \texttt{ASCB Prompt}, the worst performing prompt, still achieves higher correlation with human judgment than FKGL, the most popular traditional metric. 

\section{Statistical Significance}\label{appendix:sig-testing}
We use the William's test to calculate statistical significance of the difference in performance between each LM evaluator and traditional metric~\cite{graham-baldwin-2014-testing}. We report the p-values in \Cref{tab:signigicance-testing}. The difference in Pearson correlation between \texttt{Llama 3.3 70B}, the best performing model, the traditional metrics is statistically significant, except for DCRS and CLI. The Pearson correlation difference between the LM evaluators and FKGL, the most popular metric, is statistically significant, except \texttt{Llama 3.1 8B}. The Kendall-Tau values show that the Mistral, Mixtral, and Gemma models are statistically significant over most of the traditional metrics. This supports our suggestions from \Cref{subsec:recommendations}, in which we recommend using a combination of the best performing traditional metrics (DCRS and CLI) with LM evaluators, while discontinuing the use of FKGL. 

\input{tables/sig-testing}

\input{tables/prompts}

%% file: tables/human_annotation_distribution.tex
\begin{table}[h!]
\centering \footnotesize
\begin{tabular}{r|l|l|l|l|l}
\textbf{Score} & 5 & 4 & 3 & 2 & 1 \\ \hline
\textbf{\%}    & 37 & 29 & 17 & 10 & 7
\end{tabular}
\vspace{-2mm}
\caption{Percentage of scores assigned in human annotated dataset for reading ease.}\label{tab:human_annotation_distribution}
\vspace{-4mm}
\end{table}

%% file: tables/prompt-comparison.tex
\begin{table}[!t]
\centering
    \begin{subtable}{.49\textwidth}
        \centering \small
       \begin{tabular}{r|lll}
        Model & Simple & ASCB & Own \\ \hline
        Mistral 7B & 0.46 & 0.54 & 0.52 \\
        Mixtral 7B & 0.46 & 0.47 & 0.54 \\
        Gemma 1.1 7B & 0.55 & 0.33 & 0.54 \\
        Llama 3.1 8B & 0.54 & 0.56 & 0.45 \\
        Llama 3.3 70B & 0.59 & 0.58 & 0.56 \\\hdash
        
        Mean Corr. & 0.52 & 0.50 & 0.52
        \end{tabular}
        \vspace{-1mm}
        \caption{Pearson Correlation.}
        \label{subtab:prompt-comparison-pearson}
        \vspace{2mm}
    \end{subtable}    
    \begin{subtable}{.49\textwidth}
        \centering \small
           \begin{tabular}{r|lll}
            Model & Simple & ASCB & Own \\ \hline
            Mistral 7B & 0.32 & 0.40 & 0.44 \\
            Mixtral 7B & 0.36 & 0.41 & 0.41 \\
            Gemma 1.1 7B & 0.42 & 0.24 & 0.43 \\
            Llama 3.1 8B & 0.38 & 0.35 & 0.34 \\
            Llama 3.3 70B & 0.36 & 0.38 & 0.35 \\\hdash
            Mean Corr. & 0.37 & 0.36 & 0.39
            \end{tabular}
        \vspace{-1mm}
        \caption{Kendall-Tau Correlation.}\label{subtab:prompt-comparison-kt}
    \end{subtable} 
    \vspace{-6mm}
    \caption{Pearson and Kendall-Tau Correlation with human judgment for each prompt listed in \Cref{tab:prompts}. \texttt{Own Reasoning} prompt performs the best averaged across all models.}
    \label{tab:prompt-comparison}
    \vspace{-6mm}
\end{table}

%% file: tables/sig-testing.tex
\begin{table*}[!htb]
\centering
    \begin{subtable}{.9\textwidth}
        \centering \footnotesize
        \begin{tabular}{l|rrrrrrrr}
         & \multicolumn{1}{l}{LW} & \multicolumn{1}{l}{Spache} & \multicolumn{1}{l}{FRE} & \multicolumn{1}{l}{ARI} & \multicolumn{1}{l}{GFI} & \multicolumn{1}{l}{DCRS} & \multicolumn{1}{l}{CLI} & \multicolumn{1}{l}{FKGL} \\ \hline
        Mistral 7B & \cellcolor[HTML]{B7E1CD}6.51E-04 & \cellcolor[HTML]{B7E1CD}0.02 & \cellcolor[HTML]{B7E1CD}0.05 & \cellcolor[HTML]{B7E1CD}0.01 & \cellcolor[HTML]{B7E1CD}0.05 & 0.19 & 0.18 & \cellcolor[HTML]{B7E1CD}0.02 \\
        Mixtral 7B & \cellcolor[HTML]{B7E1CD}6.90E-04 & \cellcolor[HTML]{B7E1CD}0.01 & \cellcolor[HTML]{B7E1CD}0.03 & \cellcolor[HTML]{B7E1CD}0.01 & \cellcolor[HTML]{B7E1CD}0.04 & 0.16 & 0.15 & \cellcolor[HTML]{B7E1CD}0.02 \\
        Gemma 7B & \cellcolor[HTML]{B7E1CD}9.65E-04 & \cellcolor[HTML]{B7E1CD}0.02 & \cellcolor[HTML]{B7E1CD}0.03 & \cellcolor[HTML]{B7E1CD}0.01 & \cellcolor[HTML]{B7E1CD}0.04 & 0.17 & 0.15 & \cellcolor[HTML]{B7E1CD}0.02 \\
        Llama 3.1 8B & \cellcolor[HTML]{B7E1CD}2.00E-03 & \cellcolor[HTML]{B7E1CD}0.04 & 0.14 & \cellcolor[HTML]{B7E1CD}0.03 & 0.10 & 0.34 & 0.31 & 0.06 \\
        Llama 3.1 70B & \cellcolor[HTML]{B7E1CD}3.66E-04 & \cellcolor[HTML]{B7E1CD}0.01 & \cellcolor[HTML]{B7E1CD}0.02 & \cellcolor[HTML]{B7E1CD}0.01 & \cellcolor[HTML]{B7E1CD}0.03 & 0.14 & 0.13 & \cellcolor[HTML]{B7E1CD}0.02 \\
        % OLMo 2 13 B & \cellcolor[HTML]{B7E1CD}1.81E-04 & \cellcolor[HTML]{B7E1CD}0.01 & \cellcolor[HTML]{B7E1CD}0.02 & \cellcolor[HTML]{B7E1CD}3.75E-03 & \cellcolor[HTML]{B7E1CD}0.02 & 0.11 & 0.11 & \cellcolor[HTML]{B7E1CD}0.01
        \end{tabular}
        \vspace{-1mm}
        \caption{Pearson correlation p-values.}\label{subtab:pearson-pval}
    \end{subtable}%
    \vspace{1mm}
    \quad
    \begin{subtable}[t]{.9\textwidth}
        \centering \footnotesize
          \begin{tabular}{l|rrrrrrrr}
         & \multicolumn{1}{l}{LW} & \multicolumn{1}{l}{Spache} & \multicolumn{1}{l}{FRE} & \multicolumn{1}{l}{ARI} & \multicolumn{1}{l}{GFI} & \multicolumn{1}{l}{DCRS} & \multicolumn{1}{l}{CLI} & \multicolumn{1}{l}{FKGL} \\ \hline
        Mistral 7B & \cellcolor[HTML]{B7E1CD}0.01 & \cellcolor[HTML]{B7E1CD}0.01 & \cellcolor[HTML]{B7E1CD}0.03 & \cellcolor[HTML]{B7E1CD}0.01 & \cellcolor[HTML]{B7E1CD}0.04 & 0.13 & 0.10 & \cellcolor[HTML]{B7E1CD}0.03 \\
        Mixtral 7B & \cellcolor[HTML]{B7E1CD}0.01 & \cellcolor[HTML]{B7E1CD}0.02 & \cellcolor[HTML]{B7E1CD}0.05 & \cellcolor[HTML]{B7E1CD}0.02 & 0.06 & 0.19 & 0.14 & \cellcolor[HTML]{B7E1CD}0.04 \\
        Gemma 7B & \cellcolor[HTML]{B7E1CD}0.01 & \cellcolor[HTML]{B7E1CD}0.02 & \cellcolor[HTML]{B7E1CD}0.03 & \cellcolor[HTML]{B7E1CD}0.02 & \cellcolor[HTML]{B7E1CD}0.05 & 0.16 & 0.10 & \cellcolor[HTML]{B7E1CD}0.03 \\
        Llama 3.1 8B & \cellcolor[HTML]{B7E1CD}0.02 & 0.05 & 0.12 & \cellcolor[HTML]{B7E1CD}0.05 & 0.12 & 0.31 & 0.25 & 0.09 \\
        Llama 3.1 70B & \cellcolor[HTML]{B7E1CD}0.03 & 0.06 & 0.09 & \cellcolor[HTML]{B7E1CD}0.05 & 0.12 & 0.30 & 0.24 & 0.09 \\
        % OLMo 2 13 B & \cellcolor[HTML]{B7E1CD}3.15E-03 & \cellcolor[HTML]{B7E1CD}0.01 & \cellcolor[HTML]{B7E1CD}0.02 & \cellcolor[HTML]{B7E1CD}0.01 & \cellcolor[HTML]{B7E1CD}0.03 & 0.10 & 0.08 & \cellcolor[HTML]{B7E1CD}0.02
        \end{tabular}
        \caption{Kendall-Tau p-values.}\label{subtab:kt-pval}
    \end{subtable} 
    \vspace{-2mm}
    \caption{William's test p-values comparing the difference in performance between each LM and each traditional metric. Values that are statistically significant ($\textrm{p-value} < 0.05$), are highlighted in green.}
    \label{tab:signigicance-testing}
    \vspace{-20mm}
\end{table*}

%% file: tables/prompts.tex
\begin{table*}[]
\centering \small
\begin{tabular}{l}
% \hline
\textbf{Simple Prompt} \\ \hline
\begin{tabular}[c]{p{.9\textwidth}}
\vspace{-1mm} 
On a scale of  1 to 5, what is the reading ease of the following text? 1 indicates the text requires expert background knowledge and 5 indicates the text is readable to the general population. Assume the reader is an adult.\textnewline \textnewline \\ Format the output as follows:\textnewline\\ Score: \textless{}score\textgreater\textnewline\\ Reason: \textless{}reasoning\textgreater \textnewline \\ Text: \{SUMMARY\}
\vspace{5mm}
\end{tabular} \\ 

% \hline
\textbf{ASCB Guidelines Prompt} \\ \hline
\begin{tabular}[]{p{.9\textwidth}}
\vspace{-1mm} 
On a scale of  1 to 5, what is the reading ease of the following text? 1 indicates the text requires expert background knowledge and 5 indicates the text is readable to the general population. Characteristics of a highly readable text include:\textnewline\\ - Know your audience, and focus and organize your information for that particular audience.\textnewline\\ - Focus on the big picture. What larger problem is your work a part of? What major ideas or issues does your work address? How will your work help global understanding of some issue?\textnewline\\ - Avoid jargon. If you must use a technical term, make sure to explain it, but simplify the language.\textnewline\\ - Try to use metaphors or analogies to everyday experiences that people can relate to.\textnewline\\ - Underscore the importance of public support for exploratory research and scientific information, and the role of this information in providing the context for effective policy making.\textnewline\textnewline\\ Assume the reader is an adult. Do not use Flesch-Kincaid or other readability formulas. Use your own judgment to rate the text.\textnewline\textnewline\\ Format the output as follows:\textnewline\\ Score: \textless{}score\textgreater\textnewline\\ Reason: \textless{}reasoning\textgreater\textnewline\textnewline\\ Text: \{SUMMARY\}
\vspace{5mm}
\end{tabular} \\

% \hline
\textbf{Own Reasoning Prompt} \\ \hline
\begin{tabular}[c]{p{.9\textwidth}}
\vspace{-1mm}
On a scale of  1 to 5, what is the reading ease of the following text? 1 indicates the text requires expert background knowledge and 5 indicates the text is readable to the general population. \textnewline\\ Assume the reader is an adult. Do not use Flesch-Kincaid or other readability formulas. Use your own judgment to rate the text.\textnewline\textnewline\\ Format the output as follows:\textnewline\\ Score: \textless{}score\textgreater\textnewline\\ Reason: \textless{}reasoning\textgreater \textnewline\textnewline\\ Text: \{SUMMARY\}
% \vspace{5mm}
\end{tabular} \\
% \hline
\end{tabular}
\caption{Prompts we tested. \texttt{Own Reasoning} is the best performing prompt, as reported in~\Cref{tab:prompt-comparison}.}\label{tab:prompts}
\end{table*}